\documentclass[conference]{IEEEtran}
\IEEEoverridecommandlockouts
\usepackage{cite}
\usepackage{amsmath,amssymb,amsfonts}
\usepackage{algorithmic}
\usepackage{graphicx}
\usepackage{textcomp}
\usepackage{xcolor}
\usepackage{placeins}
\usepackage{booktabs}
\usepackage{float}
\usepackage{tabularx}
\usepackage{booktabs}
\def\BibTeX{{\rm B\kern-.05em{\sc i\kern-.025em b}\kern-.08em
    T\kern-.1667em\lower.7ex\hbox{E}\kern-.125emX}}
\begin{document}

\title{
	A Bio-Plausible Visual Neural Network for Locust-Inspired Collision Perception
	\thanks{This research was supported by the National Natural Science Foundation of China under grant nos. 62376063, 12571558.}
}

\author{
	\IEEEauthorblockN{Qinbing Fu$^{*,\dagger,1}$, Jiani Li$^{\dagger,1}$, Jiajun Huang$^{1}$, Jigen Peng$^{1}$}
	\IEEEauthorblockA{$^{1}$ School of Mathematics and Information Science, Guangzhou University, China}
	\IEEEauthorblockA{$^{*}$ Corresponding author: qifu@gzhu.edu.cn}
	\IEEEauthorblockA{$\dagger$ The authors contributed equally.}
}

\maketitle

\begin{abstract}
	
Locust visual systems have long served as an important biological paradigm for studying looming perception and collision avoidance. 
Numerous computational models have successfully reproduced the selective responses of Lobula Giant Movement Detector (LGMD) neurons to approaching objects, thereby emulating the fundamental functionality of the biological system. 
However, existing models remain limited in biological plausibility and robustness when operating in complex and dynamic visual environments. 
To address these limitations, we propose a biologically plausible neural network for locust-inspired looming detection. 
The proposed framework incorporates a spatially isotropic sampling strategy that mimics the ommatidial organization of the locust compound eye, a population-voting mechanism inspired by population coding in biological neural systems, and leaky integrate-and-fire neuronal dynamics to replace conventional sigmoid-based membrane activation. 
Systematic experiments on synthetic stimuli, laboratory sequences, and real-world driving scenarios demonstrate that the proposed model improves robustness under challenging visual conditions while preserving computational efficiency and enhancing biological fidelity. 
These results highlight the potential of biologically grounded neural computation for robust and efficient collision perception.

\end{abstract}


\section{Introduction}
\label{Sec: introduction}

Looming detection is the ability of a visual system to perceive an approaching object through the continuous expansion of its projected image size over time \cite{b1,b2}. 
Unlike static object recognition, it relies on temporal expansion cues and is therefore well suited to time-critical perception tasks \cite{b2}. 
This capability supports collision perception and rapid avoidance behaviors in both biological organisms and artificial systems \cite{b2,b7}. 
Accordingly, reliable looming detection is essential for real-time safety decisions in autonomous navigation applications, including robotics and autonomous driving \cite{b1,b5}.

Traditional visual approaches to motion perception and collision detection are primarily based on object segmentation, optical flow estimation, and deep learning. 
Segmentation methods rely on accurate foreground extraction, whereas optical flow techniques estimate pixel-wise motion fields to infer object trajectories. 
Deep learning approaches further improve detection performance by learning hierarchical visual representations from large-scale annotated datasets. 
Despite their success, these methods generally incur substantial computational and memory costs and often require extensive training data, limiting their applicability to real-time, resource-constrained systems.

To overcome these limitations, bio-inspired and biologically plausible visual neural systems have attracted increasing attention. 
The locust visual system provides a well-established biological model for looming detection, in which Lobula Giant Movement Detector (LGMD) neurons respond strongly to approaching stimuli while suppressing translational and receding motion \cite{b2,b3}. 
Classical LGMD-based computational models reproduce these properties through synaptic integration and neural filtering mechanisms \cite{b3,b5}. 
In parallel, direction-selective pathways, such as the fly-inspired LPLC2 system, further enhance motion encoding by incorporating spatial and directional selectivity \cite{b8}.

Despite their biological interpretability and computational efficiency, existing bio-inspired models still face several limitations. 
Most LGMD-based frameworks rely on a single sigmoid output unit for collision decision-making, making them vulnerable to noise, background clutter, and local motion disturbances \cite{b3,b5}. 
Moreover, the lack of a biologically plausible mechanism for integrating distributed spatial responses further weakens their robustness in real-world environments. 
In addition, fixed pixel-wise receptive-field structures deviate from the ommatidial organization of the locust compound eye, which may limit adaptability in complex visual scenes. 
More broadly, current LGMD models have not yet fully exploited the population-level computation principles observed in biological vision systems \cite{b9,b10,b11}.

In this work, we propose a biologically plausible visual neural network for locust-inspired looming detection. 
Unlike previous LGMD-based models, the proposed framework incorporates a spatial sampling strategy that mimics the ommatidial organization of the locust compound eye, enabling biologically faithful spatial encoding \cite{b7}. 
To improve robustness, a population-voting mechanism is introduced to integrate responses from multiple local visual sub-fields, inspired by population coding principles observed in biological neural systems \cite{b9,b10,b11}. 
Furthermore, self-inhibition and lateral inhibition mechanisms are incorporated to enhance looming selectivity by suppressing responses to translational motion \cite{b4}. 
Finally, the output neuron is modeled as a leaky integrate-and-fire (LIF) neuron, providing a biologically plausible representation of the temporal dynamics of neuronal spiking \cite{b12,b13}.

Extensive experiments were conducted using synthetic stimuli, laboratory sequences, and real-world ground-vehicle scenes, covering both approaching and translational motions embedded in complex dynamic environments. 
In particular, we visualized the spatiotemporal interactions between excitatory and inhibitory neuronal dynamics to illustrate the effects of self-inhibition and lateral inhibition. 
The proposed model was further compared with classical LGMD-based methods and biological LGMD neuronal dynamics, demonstrating improved biological plausibility, robustness, and stability while maintaining low computational complexity.

\section{Method}
\label{Sec: method}

Generally, the proposed neural network is designed to emulate the hierarchical processing strategy of the locust's optic lobe, transforming dynamic visual inputs into collision-sensitive neural responses through multiple biologically inspired stages.

\begin{figure}[t]
	\vspace{-20pt}
	\centering
	\includegraphics[width=\linewidth]{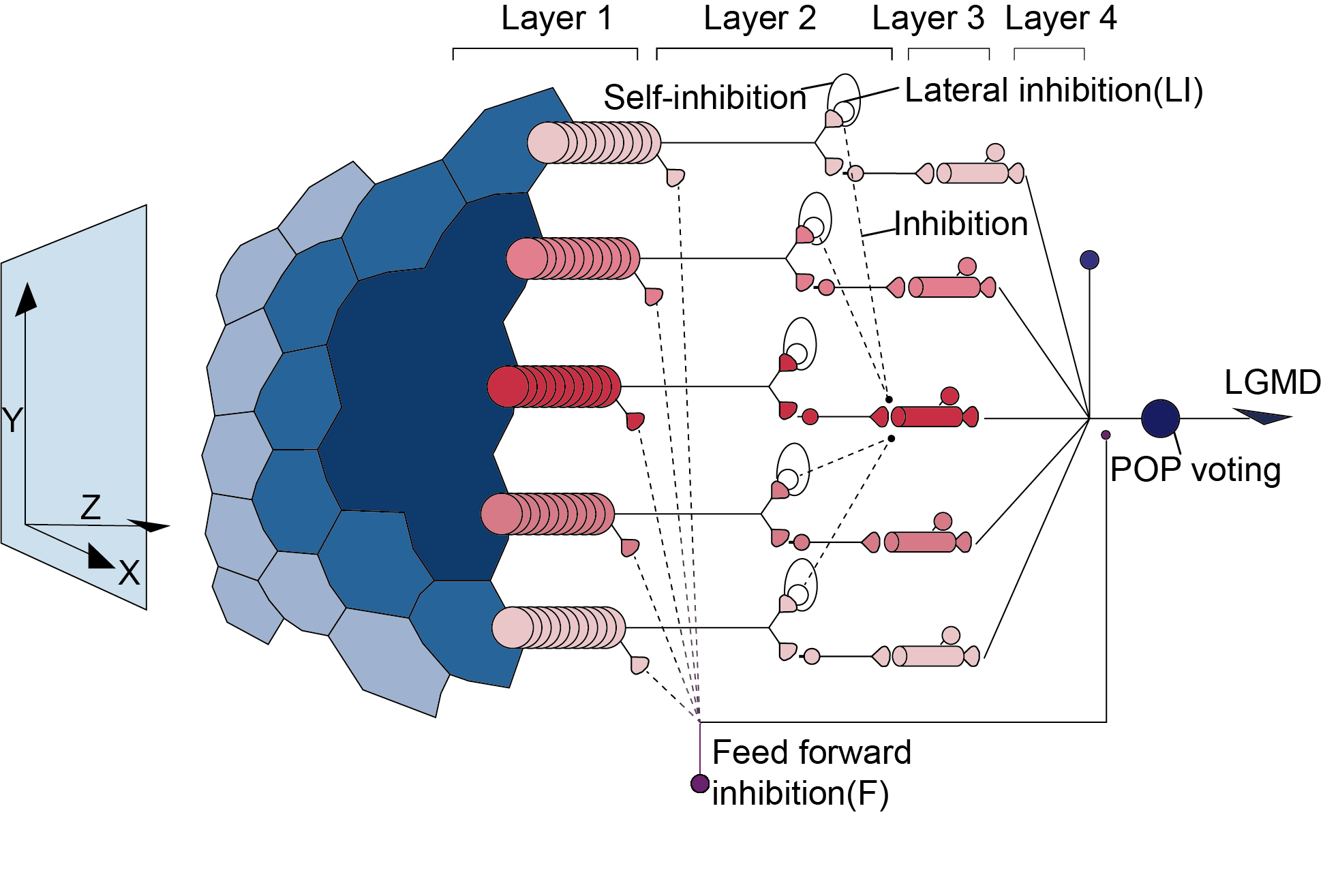}
	\caption{
		Overall architecture of the proposed neural network, incorporating ommatidial spatial sampling, coordinated inhibitory processing, and a population-voting mechanism, adapted from \cite{b3}.
		}
	\vspace{-10pt}
	\label{fig1}
\end{figure}

As shown in Fig.~\ref{fig1}, the model first performs spatial encoding using a hexagonal ommatidial sampling lattice, which preserves isotropic spatial relationships and mimics the organization of the compound eye. 
The sampled signals are then processed through ON/OFF motion channels to extract temporally varying luminance changes. 
These motion signals are further shaped by excitatory (E), feed-forward inhibitory (FFI), lateral inhibitory (LI), and self-inhibitory (SI) interactions, forming a distributed neural representation of motion saliency. 
A population-voting mechanism is subsequently applied to integrate spatially distributed responses, thereby improving robustness against local noise and motion ambiguity. 
Finally, the aggregated population response is encoded through leaky integrate-and-fire (LIF) neuronal dynamics.

\subsection{Isotropic Spatial Sampling}

\begin{figure}[t]
	\vspace{-20pt}
	\centering
	\includegraphics[width=0.8\linewidth]{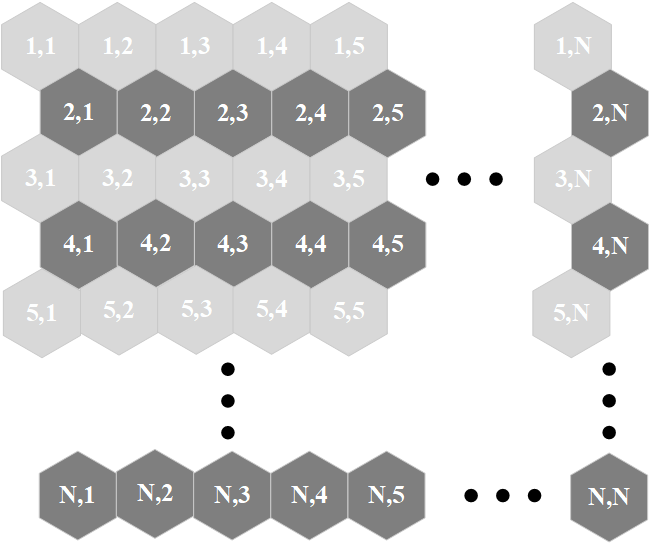}
	\caption{
		Isotropic spatial sampling of the hexagonal ommatidia placement mimicking the locust compound eye.
		}
	\vspace{-10pt}
	\label{fig2}
\end{figure}

To emulate the compound eye of the locust, the input image $L(x,y,t)$ is projected onto a hexagonal lattice. 
Each unit corresponds to a local receptive field $H_{\mathrm{rc}}$. 
The center of each hexagonal grid is defined as $(x_{\mathrm{rc}}, y_{\mathrm{rc}})$, where $r$ and $c$ denote row and column indices, respectively. 
As shown in Fig.~\ref{fig2}, the hexagonal sampling structure preserves isotropic spatial relationships and reduces directional bias compared to rectangular grids. 
Each hexagonal unit outputs the mean intensity of all pixels within its receptive field as
\begin{equation}
\hat{L}_{\mathrm{rc}}(t)=\frac{1}{|H_{\mathrm{rc}}|}\sum_{(x,y)\in H_{\mathrm{rc}}} L(x,y,t).
\end{equation}
The hexagonal lattice is constructed using a staggered row-shift geometry, where odd rows are horizontally offset by half a grid spacing relative to even rows. 
The corresponding center coordinates are defined as
\begin{equation}
x_{rc} = \left(c + 0.5 + 0.5\,\mathrm{mod}(r,2)\right)\Delta x,\quad
y_{rc} = (r+0.5)\Delta y,
\end{equation}
where $\Delta x=\sqrt{3}r_0$ and $\Delta y=1.5r_0$.

\subsection{Multi-Layer Processing of ON/OFF-Contrast}

After spatial sampling, motion signals are first transmitted to the photoreceptor layer. 
Temporal derivatives are computed to capture ``change" of brightness as
\begin{equation}
P(x,y,t)=\hat{L}_{\mathrm{rc}}(t)-\hat{L}(x,y,t-\Delta t).
\end{equation}
To simulate neural diffusion in the optical layer, a Gaussian smoothing operation is applied as
\begin{equation}
\tilde{P}(x,y,t)=\mathcal{B} * P(x,y,t),
\end{equation}
where $*$ defines a convolution process, and $\mathcal{B}$ is given by
\begin{equation}
\mathcal{B}=
\begin{bmatrix}
0.06 & 0.12 & 0.06\\
0.12 & 0.28 & 0.12\\
0.06 & 0.12 & 0.06
\end{bmatrix}.
\end{equation}
The output is subsequently fed into an FFI pathway, which suppresses excessive global activation when a large number of photoreceptors are simultaneously excited. 
The FFI signal is defined as the spatial average of photoreceptor activation as
\begin{equation}
FFI(t)=\frac{1}{R \cdot C}\sum_{x=1}^{R}\sum_{y=1}^{C} \tilde{P}(x,y,t-\Delta t).
\end{equation}
The FFI tunes a time-varying inhibitory modulation weight as
\begin{equation}
\omega(t)=\max\left(0.3,\frac{FFI(t)}{T_{\mathrm{ffi}}}\right),
\end{equation}
where $T_{ffi}$ is a normalization constant controlling inhibition sensitivity with the baseline weight at 0.3.

Subsequently, motion signals are split into ON and OFF polarity channels where the ON-contrast signal is obtained by
\begin{equation}
ON(x,y,t)=
\begin{cases}
\tilde{P}(x,y,t) + \lambda \cdot ON(x,y,t-\Delta t), & \tilde{P}>\mathcal{C}_{\mathrm{on}}, \\
\lambda \cdot ON(x,y,t-\Delta t), & \text{otherwise},
\end{cases}
\end{equation}
whilst the OFF-contrast signal is computed as
\begin{equation}
OFF(x,y,t)=
\begin{cases}
|\tilde{P}(x,y,t)| + \lambda \cdot OFF(x,y,t-\Delta t), & \tilde{P}<\mathcal{C}_{\mathrm{off}}, \\
\lambda \cdot OFF(x,y,t-\Delta t), & \text{otherwise},
\end{cases}
\end{equation}
where $\lambda$ is a temporal decay factor, and $\mathcal{C}_{\mathrm{on}}$ and $\mathcal{C}_{\mathrm{off}}$ are clipping-point threshold controlling activation sensitivity.

To model population-level activity, ON/OFF responses are aggregated across the spatial field:
\begin{equation}
ON_\mathrm{s}(t) = \max \left( T_{\mathrm{on}}, \frac{1}{R \cdot C} \sum_{x=1}^{R} \sum_{y=1}^{C} ON(x,y,t) \right),
\end{equation}
\begin{equation}
OFF_\mathrm{s}(t) = \max \left( T_{\mathrm{off}}, \frac{1}{R \cdot C} \sum_{x=1}^{R} \sum_{y=1}^{C} OFF(x,y,t) \right),
\end{equation}
where $T_{\mathrm{on}}$ and $T_{\mathrm{off}}$ denote activation thresholds for the ON and OFF pathways, respectively.

These ON/OFF-contrast signals contribute to downstream processing layers enabling population-based response modulation. 
We take the ON-pathway neural processing to articulate the remaining computation where the OFF-pathway follows an identical computational structure.

Local excitation is computed by neighborhood aggregation:
\begin{equation}
E_{\mathrm{on}}(x,y,t) = ON(x,y,t) * W_{\mathrm{e}},
\end{equation}
where $W_{\mathrm{e}}$ is a Gaussian-distributed kernel defining the excitatory receptive field. 
Lateral inhibition refers to spatial competition between neighboring neurons, where the activation of one neuron suppresses the responses of surrounding neurons. 
It is implemented as a convolution over the excitation map:
\begin{equation}
LI_{\mathrm{on}}(x,y,t) = E_{\mathrm{on}}(x,y,t-\Delta t) * W_{\mathrm{l}}.
\end{equation}
In addition, self-inhibition refers to activity-dependent suppression within a local region of the same neuron population. 
It reduces excessive local excitation by modulating responses based on the local visual field (LVF) as
\begin{equation}
LVF_{\mathrm{on}}(r,c,t) =
\frac{1}{9}
\sum_{i=-1}^{1}\sum_{j=-1}^{1}
E_{\mathrm{on}}(3r+i,3c+j,t-\Delta t),
\end{equation}
and the self-inhibition response is defined as
\begin{equation}
SI_{\mathrm{on}}(x,y,t)=
\begin{cases}
\mathcal{I}_{\mathrm{on}}(x,y,t), &LVF_{\mathrm{on}} \le \gamma, \\
\beta \cdot \mathcal{I}_{\mathrm{on}}(x,y,t), &\text{otherwise},
\end{cases}
\end{equation}
where
\begin{equation}
\mathcal{I}_{\mathrm{on}}(x,y,t)= E_{\mathrm{on}}(x,y,t-\Delta t) * W_{\mathrm{s}},
\end{equation}
where $\gamma$ and $\beta$ control local-field activation and SI strength, respectively. 
Moreover, the LI kernel is defined as
\begin{equation}
W_{\mathrm{l}} =
\begin{bmatrix}
0.01 & 0.02 & 0.03 & 0.02 & 0.01 \\
0.02 & 0.06 & 0.10 & 0.06 & 0.02 \\
0.03 & 0.10 & 0.16 & 0.10 & 0.03 \\
0.02 & 0.06 & 0.10 & 0.06 & 0.02 \\
0.01 & 0.02 & 0.03 & 0.02 & 0.01
\end{bmatrix},
\end{equation}
and the SI is defined as
\begin{equation}
W_{\mathrm{s}} = 2 \cdot W_{\mathrm{l}}.
\end{equation}

Subsequently, there are ON/OFF-summation layers combining E, LI, and SI signals. 
That is,
\begin{equation}
\mathcal{S}_{\mathrm{on}}(x,y,t)=\max\big(0, S_{\mathrm{on}}(x,y,t)\big),
\end{equation}
where,
\begin{equation}
S_{\mathrm{on}}(x,y,t)=E_{\mathrm{on}}(x,y,t) - \theta \cdot SI_{\mathrm{on}}(x,y,t) - \omega(t) \cdot LI_{\mathrm{on}}(x,y,t).
\end{equation}
The OFF-summation layer computation is consistent.

\subsection{Population-Voting Mechanism}

Unlike previous modeling studies, we introduce a population-voting mechanism that integrates local neural responses across neighboring receptive fields to generate a robust global response. 
That is, 
\begin{equation}
POP_{\mathrm{on}}(x,y,t) = \mathcal{S}_{\mathrm{on}}(x,y,t) * W_{\mathrm{pop}},
\end{equation}
\begin{equation}
POP_{\mathrm{off}}(x,y,t) = \mathcal{S}_{\mathrm{off}}(x,y,t) * W_{\mathrm{pop}},
\end{equation}
where $W_{\mathrm{pop}}$ is a spatial uniform-distributed 11 $\times$ 11 kernel. 
The binary activation maps are then computed as
\begin{equation}
BIN_{\mathrm{on}}(x,y,t)
=
\begin{cases}
1, & POP_{\mathrm{on}}(x,y,t) \ge \alpha ON_\mathrm{s}(t), \\
0, & \text{otherwise},
\end{cases}
\end{equation}
\begin{equation}
BIN_{\mathrm{off}}(x,y,t)
=
\begin{cases}
1, & POP_{\mathrm{off}}(x,y,t) \ge \alpha OFF_\mathrm{s}(t), \\
0, & \text{otherwise},
\end{cases}
\end{equation}
\begin{equation}
S(x,y,t)
=
\begin{cases}
1, & Bin_{\mathrm{on}}(x,y,t) + Bin_{\mathrm{off}}(x,y,t) \ge 1, \\
0, & \text{otherwise}.
\end{cases}
\end{equation}
This computational mechanism mimics population-level spike activation, in which only sufficiently strong local responses contribute to the downstream decision-making neuron.

The population activity is integrated across the whole receptive field as
\begin{equation}
\mathcal{K}(t) = \sum_{x=1}^{R} \sum_{y=1}^{C} S(x,y,t).
\end{equation}

\subsection{Neuronal Dynamics}

\begin{figure}[t]
    \centering
    \vspace{-20pt}
    \includegraphics[width=0.9\linewidth]{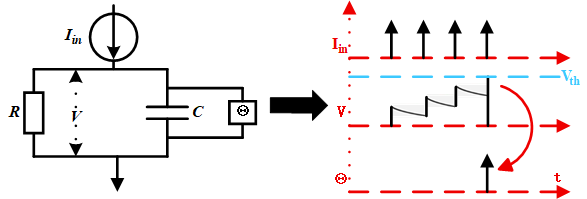}
    \caption{
    	Leaky integrate-and-fire (LIF) neuronal circuit and its membrane potential dynamics.
    	}
    \vspace{-10pt}
    \label{fig3}
\end{figure}

The final collision decision is generated by a LIF neuron, which models the temporal integration dynamics of biological LGMD neurons. 
The membrane potential evolves according to
\begin{equation}
\tau_{\mathrm{m}} \frac{dV(t)}{dt} = -(V(t) - V_{\mathrm{rest}}) + R_\mathrm{m} I(t).
\end{equation}
When the membrane potential reaches the firing threshold, a spike is generated and the potential is reset. 
That is,
\begin{equation}
\text{if } V(t) \ge V_{\mathrm{th}}, \quad V(t) \leftarrow V_{\mathrm{reset}}.
\end{equation}
The input current is defined based on population activity dynamics:
\begin{equation}
I(t)=\max\left(0, \mathcal{K}(t)+\frac{(\mathcal{K}(t)-\mathcal{K}(t-\Delta t)) \cdot f_{\mathrm{ps}}}{0.8} \right),
\end{equation}
where $f_{\mathrm{ps}}$ indicates frames per second in discrete sampling of video sequences. 
This mechanism captures the key electrophysiological characteristics of LGMD neurons, including temporal integration, threshold-triggered spike generation, and post-spike reset dynamics. 
Table~\ref{tab:bio_parameters} lists the key parameters participating in neural computation.

\begin{table}[!h] 
\centering 
\caption{The Parameters Setup} 
\begin{tabular}{ll} \toprule \textbf{Parameter} & \textbf{Description}\\
\midrule 
$T_{\mathrm{ffi}} \in [8,12]$ & FFI coefficient \\ 
$\mathcal{C}_{\mathrm{on}}=0.1$ & ON-pathway clipping-point threshold \\ 
$\mathcal{C}_{\mathrm{off}}=-0.1$ & OFF-pathway clipping-point threshold \\ 
$T_{\mathrm{on}}, T_{\mathrm{off}} \in [0.1,0.5]$ & Activation threshold of ON/OFF pathway \\
$\theta \in [0.2,3]$ & SI bias \\ 
$\gamma \in (0, 3]$ & LVF threshold \\ 
$\beta \in [0.1,0.5]$ & SI coefficient \\ 
$\alpha \in [2,7]$ & Voting threshold coefficient \\
$R=100, C=100$ & rows and columns of spatial sampling \\ 
\bottomrule 
\end{tabular} 
\label{tab:bio_parameters} 
\end{table}

\section{Results and Analysis}
\label{Sec: results}

This section presents the experimental results and corresponding analyses. 
The visual stimuli used to evaluate the proposed biologically plausible model are categorized into two groups: controlled laboratory video sequences and real-world dashboard-camera recordings containing both collision and non-collision driving scenarios. 
We first evaluate the fundamental functionality and motion selectivity of the proposed model, then investigate the contributions of its key neural mechanisms, and finally assess its robustness using a large-scale dataset comprising hundreds of vehicle video clips, adapted from \cite{b14}. 
All experiments were conducted on a standard workstation equipped with an Intel Core i5-10500 CPU and 16~GB of RAM.


\begin{figure}[t]
	\centering
	\vspace{-20pt}
	\includegraphics[width=\linewidth]{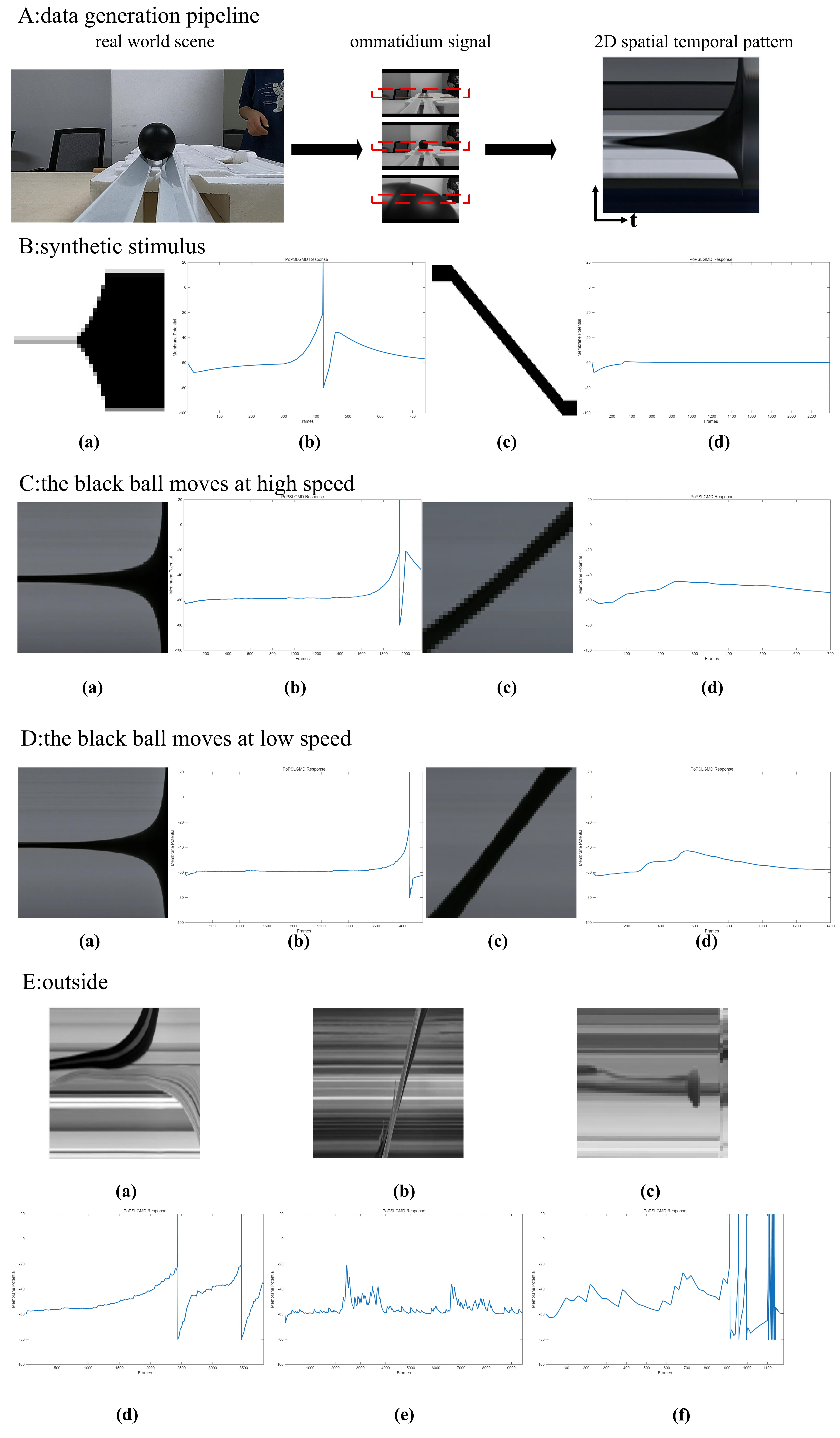}
	\caption{
		Illustration of the model's selectivity to approaching objects under diverse real-world stimuli, including looming and translational motion of an indoor ball and an outdoor ground vehicle.
	}
	\label{fig8}
	\vspace{-10pt}
\end{figure}

\subsection{Basic Functionality and Selectivity}

As shown in Fig.~\ref{fig8}, the visual stimuli are represented as one-dimensional motion signals, clearly illustrating the temporal characteristics of looming and translational motion. 
In the synthetic experiments, an expanding black square was used to simulate looming motion, whereas a translating black bar served as the translational stimulus. 
The proposed neural network responded selectively to looming motion by generating action potentials while remaining largely inactive during translational motion.

We further evaluated the model using real laboratory sequences containing high- and low-speed approaching and translating ball motions. 
In both cases, the proposed model responded selectively to approaching motion regardless of the target speed. 
Finally, the model was tested on real-world dashboard-camera recordings containing collision and non-collision driving scenarios. 
It robustly detected collision cues in crash sequences while remaining largely inactive during non-collision scenarios. 
Collectively, these experiments demonstrate the fundamental collision-perception capability and strong looming selectivity of the proposed model across a wide range of visual stimuli, from controlled synthetic sequences to complex real-world environments.


\subsection{Model Investigations}

\subsubsection{Spatial Sampling}

To emulate the hexagonal arrangement of ommatidia in the locust compound eye, the input layer employs a hexagonal grid-based spatial sampling strategy. 
Unlike conventional rectangular downsampling, this approach preserves more isotropic spatial relationships between neighboring sampling units, consistent with the biological organization of ommatidia. 
Each sampling unit represents a local receptive field that aggregates photoreceptor responses within its surrounding region.

To quantitatively assess the effectiveness of this sampling strategy, we compare it with bilinear interpolation and Gaussian downsampling under identical experimental conditions. 
As shown in Table~\ref{tab:downsample}, the proposed hexagonal scheme provides better edge preservation and structural consistency while maintaining comparable information entropy. 
These results suggest that the biologically plausible spatial encoding is better suited to edge-sensitive motion-processing tasks. 
By improving the uniformity of spatial coverage and reducing directional bias, the proposed design enhances the robustness of motion perception in complex and dynamic scenes.

\begin{table}[t]
\centering
\vspace{-20pt}
\caption{Comparison of Spatial Sampling Methods}
\label{tab:downsample}
\renewcommand{\arraystretch}{1.2}
\setlength{\tabcolsep}{4pt}
\begin{tabular}{lccc}
\toprule
Method & Entropy (bits) & Mean-Gradient & HOG-Similarity \\
\midrule
The proposed sampling  & 7.793 & \textbf{0.384} & 0.958 \\
Bilinear interpolation & 7.793 & 0.311 & 0.997 \\
Gaussian down-sampling & 7.793 & 0.295 & 0.982 \\
\bottomrule
\end{tabular}
\vspace{-10pt}
\end{table}

\subsubsection{Visualization of Spatiotemporal Signal Patterns}

To assess the biological plausibility of the proposed model, we analyze the dynamics of the excitatory, self-inhibitory, lateral-inhibitory, and summation layers. 
Their intermediate spatiotemporal response patterns are visualized using ``bubble maps", in which bubble size represents signal intensity with respect to time.

\begin{figure}[t]
	\centering
	\vspace{-20pt}
	\includegraphics[width=0.90\linewidth]{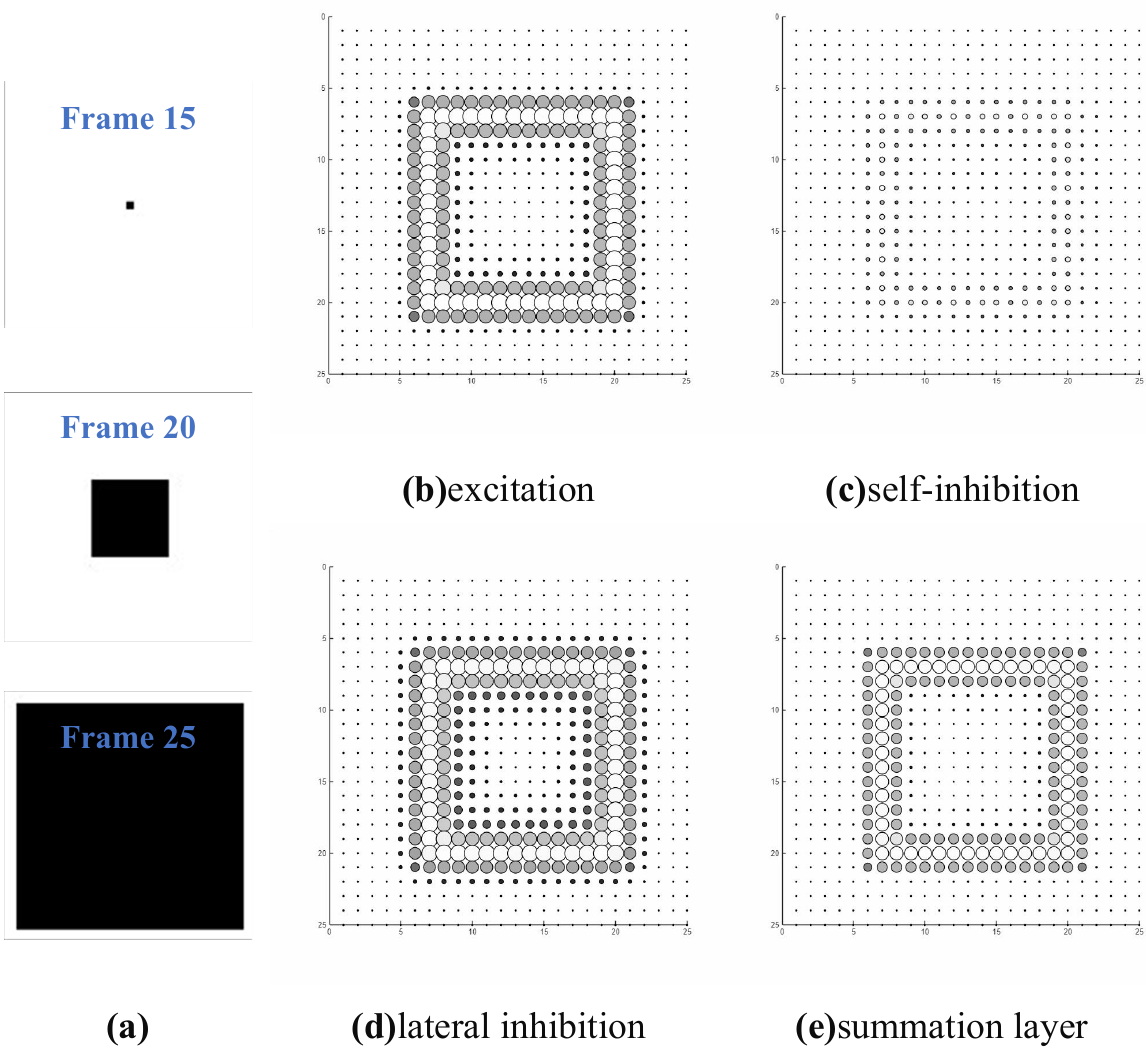}
	\caption{
		Visualization of intermediate spatiotemporal neural responses of the proposed model under looming stimulation at frame-25. 
		(a) Looming stimulus; (b) excitation; (c) self-inhibition; (d) lateral inhibition; and (e) residual excitation after inhibitory modulation. 
		The area of each bubble represents the corresponding signal intensity.
	}
	\label{fig4}
\end{figure}

For looming stimuli (Fig.~\ref{fig4}), a black square approaching along a direct collision trajectory is used as the visual input. 
During the early phase of approach, excitatory responses are weak and spatially dispersed across the receptive field. 
As the object draws closer, the excitation gradually strengthens and becomes more spatially concentrated, reflecting the increasing saliency of the looming stimulus.

Although both SI and LI also increase over time, their responses lag behind the excitation. 
Consequently, the summation layer remains primarily excitation-driven during the early stage of approach. 
As the stimulus becomes more imminent, LI becomes increasingly dominant under strong local activation, enhancing spatial competition and suppressing redundant responses. 
Overall, the summation activity increases progressively over time, producing a response pattern consistent with looming selectivity.

\begin{figure}[t]
	\centering
	\includegraphics[width=0.90\linewidth]{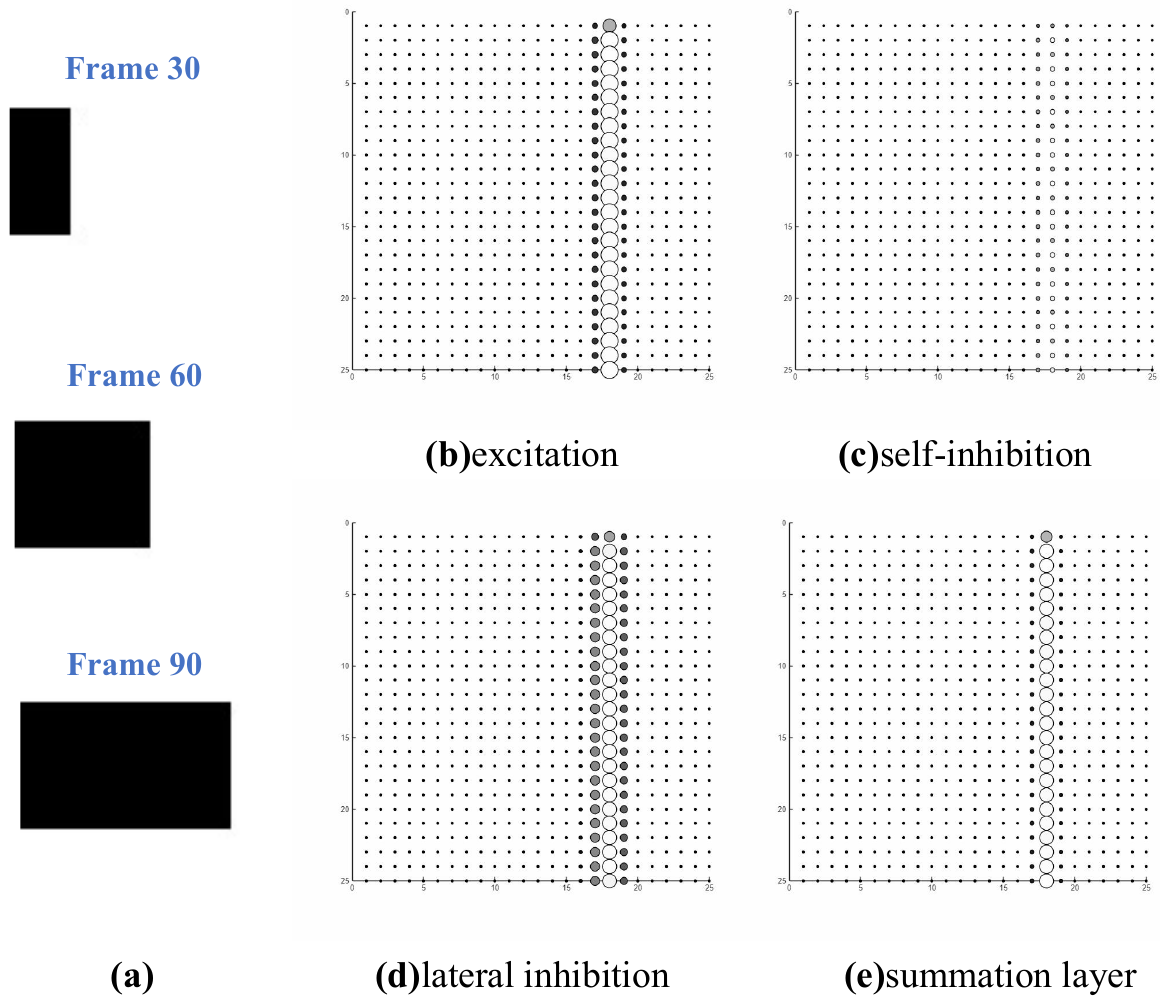}
	\caption{
		Visualization of intermediate spatiotemporal neural responses of the proposed model under translational stimulation at frame-90.
	}
	\label{fig5}
	\vspace{-10pt}
\end{figure}

For translational motion (Fig.~\ref{fig5}), a black bar moves laterally across the visual field at a constant velocity. 
In this case, the excitatory responses remain weak and spatially localized throughout the sequence. 
Although a brief transient response appears at motion onset, the excitation rapidly decays and remains low for the rest of the stimulus.

By contrast, SI and LI maintain relatively stable activity throughout the sequence. 
Their combined effects in the summation layer suppress the weak excitatory inputs, resulting in consistently low output activity. 
Consequently, translational motion does not produce substantial response accumulation, demonstrating the model's effective suppression of non-looming stimuli.

\subsubsection{Robustness Against Noise}

\begin{figure}[t]
	\vspace{-20pt}
    \centering
    \includegraphics[width=\linewidth]{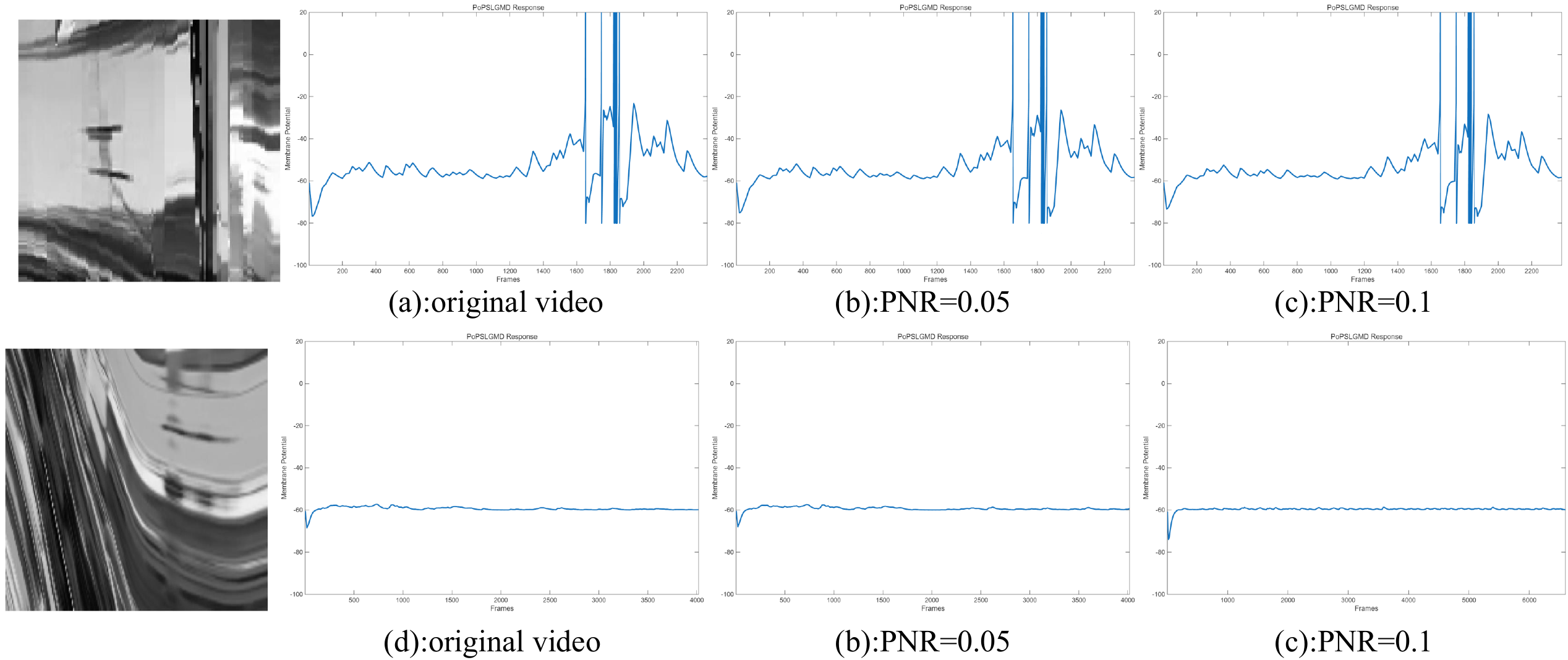}
    \caption{
    	Robustness evaluation of the proposed model under Gaussian noise conditions. 
    	The first- and second-row panels indicate results under looming and translating motion, respectively.
    	}
    \label{fig6}
    \vspace{-10pt}
\end{figure}

The proposed population-voting mechanism is expected to reduce the adverse effects of visual noise. 
To test this hypothesis, Gaussian noise with different variance levels was added to both collision and translational motion sequences. 
As shown in Fig.~\ref{fig6}, the model maintains stable response characteristics under these noise perturbations. 
This robustness arises because the population-voting mechanism effectively performs spatial averaging, thereby reducing the influence of zero-mean stochastic noise. 
In addition, the redundancy provided by population coding helps prevent isolated noise-induced excitations from dominating the final response.

\subsubsection{Neuronal dynamics}

\begin{figure}[!h]
    \centering
    \includegraphics[width=\linewidth]{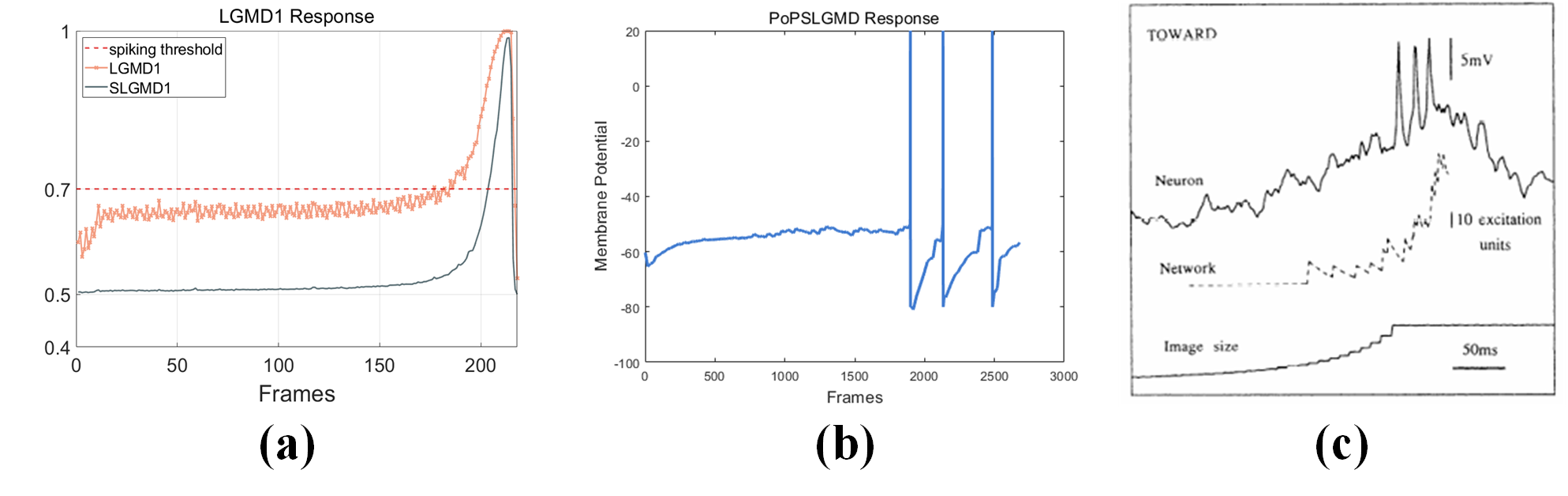}
    \caption{
    	Comparison of the responses of (a) a previous LGMD model and (b) the proposed model to looming stimuli, with reference to (c) the response of a biological LGMD neuron.
    	}
    \label{fig7}
    \vspace{-10pt}
\end{figure}

To improve biological plausibility, the resulting global signal is directly fed into an LIF neuron driving membrane potential dynamics without additional nonlinear transformations. 
As shown in Fig.~\ref{fig7}, this preserves temporal accumulation characteristics and generates event-driven spiking responses consistent with biological neurons.

\subsection{Robustness in Vehicle Data Set}

Finally, to comprehensively evaluate the robustness of the proposed model, we tested it on 194 real-world dashcam sequences, including 42 collision cases and 152 non-collision cases. 
The results are summarized in Table~\ref{tab:cm_194}. 
The proposed model achieves an accuracy of 76.29\%, a recall of 83.33\%, and an F1-score of 60.35\%, where the F1-score represents the harmonic mean of precision and recall. 
These results indicate that the model is highly sensitive to collision events.

Because the dataset is imbalanced, with non-collision cases accounting for the majority of samples, the high recall is particularly important, as it shows that most true collision events are successfully identified. 
This property is critical for safety-oriented applications, where missed detections may have severe consequences. 
However, the precision of 47.30\% indicates that false positives still occur, mainly due to complex real-world motion patterns, background motion, and camera-induced disturbances. 
Despite this limitation, the model maintains an F1-score of 60.35\%, reflecting a reasonable balance between collision sensitivity and detection precision.


\section{Summary}
\label{Sec: conclusion}

This paper presents a biologically plausible neural network for locust-inspired looming perception that integrates spatially isotropic sampling, population voting, and spiking neural dynamics into a unified computational framework for collision detection. 
Unlike conventional LGMD-based models, the proposed framework exploits distributed population coding to enhance robustness against complex visual disturbances while preserving looming selectivity and improving biological fidelity.

\begin{table}[t]
	\centering
	\vspace{-20pt}
	\caption{Performance on dashcam recording sequences}
	\begin{tabular}{cccccccc}
		\hline
		TP & FP & FN & TN & Accuracy & Precision & \textbf{Recall} & F1-score \\
		\hline
		35 & 39 & 7 & 113 & 76.29\% & 47.30\% & \textbf{83.33\%} & 60.35\% \\
		\hline
	\end{tabular}
	\label{tab:cm_194}
	\vspace{-10pt}
\end{table}

\end{document}